\documentclass[10pt,twocolumn,letterpaper]{article}

\usepackage{style}
\usepackage{times}
\usepackage{epsfig}
\usepackage{graphicx}
\usepackage{amsmath}
\usepackage{amssymb}

\usepackage{array}
\usepackage{xcolor}

\newcommand*{\rowstyle}[1]{
	\gdef\@rowstyle{#1}
	\@rowstyle\ignorespaces%
}
\newcolumntype{=}{
	>{\gdef\@rowstyle{}}
}
\newcolumntype{+}{
	>{\@rowstyle}
}

\usepackage{caption}
\usepackage{subcaption} 

\usepackage[pagebackref=true,breaklinks=true,letterpaper=true,colorlinks,bookmarks=false]{hyperref}

\PLfinalcopy

\usepackage{amssymb}
\usepackage{amsmath,amsfonts}
\usepackage{amsopn}
\usepackage{bm} 
\usepackage{multirow}

\newcommand{\vct}[1]{\boldsymbol{#1}} 

\newcommand{\ProbOpr}[1]{\mathbb{#1}}

\newcommand{\expect}[2]{
\ifthenelse{\equal{#2}{}}{\ProbOpr{E}_{#1}}
{\ifthenelse{\equal{#1}{}}{\ProbOpr{E}\left[#2\right]}{\ProbOpr{E}_{#1}\left[#2\right]}}}
\newcommand{\var}[2]{
\ifthenelse{\equal{#2}{}}{\ProbOpr{VAR}_{#1}}
{\ifthenelse{\equal{#1}{}}{\ProbOpr{VAR}\left[#2\right]}{\ProbOpr{VAR}_{#1}\left[#2\right]}}}

\newcommand{\vp}{\vct{p}}

\newcommand{\vw}{\vct{w}}

\newcommand{\eat}[1]{}
\newcommand{\method}[1]{\textsc{#1}}

\newcommand{\APBEV}{AP$_\text{BEV}$\xspace}
\newcommand{\AP}{AP$_\text{3D}$\xspace}
\newcommand{\AVOD}{\method{AVOD}\xspace}
\newcommand{\AVODC}{\method{AVOD}\xspace}
\newcommand{\AVODGT}{\method{AVOD}\xspace}
\newcommand{\Frustum}{\method{F-PointNet}\xspace}
\newcommand{\Mono}{\method{Mono3D}\xspace}
\newcommand{\DOP}{\method{3DOP}\xspace}
\newcommand{\DORN}{\method{DORN}\xspace}
\newcommand{\MLF}{\method{MLF}\xspace}
\newcommand{\MLFmono}{\method{MLF-mono}\xspace}
\newcommand{\MLFstereo}{\method{MLF-stereo}\xspace}
\newcommand{\PSMNet}{\method{PSMNet}\xspace}
\newcommand{\PSMNetpd}{\method{PSMNet$\star$}\xspace}
\newcommand{\DispNet}{\method{DispNet}\xspace}
\newcommand{\DispNetS}{\method{DispNet-S}\xspace}
\newcommand{\DispNetC}{\method{DispNet-C}\xspace}
\newcommand{\SPSstereo}{\method{SPS-stereo}\xspace}

\ifPLfinal\fi
\begin{document}

\title{Pseudo-LiDAR from Visual Depth Estimation:\\
Bridging the Gap in 3D Object Detection for Autonomous Driving}

\author{Yan Wang, Wei-Lun Chao, Divyansh Garg, Bharath Hariharan, Mark Campbell, and Kilian Q. Weinberger\\
Cornell University, Ithaca, NY\\
{\tt\small \{yw763, wc635, dg595, bh497, mc288, kqw4\}@cornell.edu}
}

\maketitle

\begin{abstract}
3D object detection is an essential task in autonomous driving. Recent techniques excel with highly accurate detection rates, provided the 3D input data is obtained from precise but expensive LiDAR technology.
Approaches based on cheaper monocular or stereo imagery data have, until now, resulted in drastically lower accuracies --- a gap that is commonly attributed to poor image-based depth estimation. 
However, in this paper we argue that it is not the quality of the data but its representation that accounts for the majority of the difference. 
Taking the inner workings of convolutional neural networks into consideration, we propose to convert image-based depth maps to pseudo-LiDAR representations  --- essentially mimicking the LiDAR signal.
With this representation we can apply different existing LiDAR-based detection algorithms. On the popular KITTI benchmark, our approach achieves impressive improvements over the existing state-of-the-art in image-based performance --- 
raising the detection accuracy of objects within the 30m range from the previous state-of-the-art of 22\% to an unprecedented 74\%. At the time of submission our algorithm holds the highest entry on the KITTI 3D object detection leaderboard for stereo-image-based approaches. Our code is publicly available at \url{https://github.com/mileyan/pseudo_lidar}.
\end{abstract}
\section{Introduction}
\label{intro}

Reliable and robust 3D object detection is one of the fundamental requirements for autonomous driving. After all, in order to avoid collisions with pedestrians, cyclist, and cars, a vehicle must be able to detect them in the first place. 

Existing algorithms largely rely on LiDAR (Light
Detection And Ranging), which provide accurate 3D point clouds of the surrounding environment. Although highly precise, alternatives to LiDAR are desirable for multiple reasons. 
First, LiDAR is expensive, which incurs a hefty premium for autonomous driving hardware.  Second, over-reliance on a single sensor is an inherent safety risk and it would be advantageous to have a secondary sensor to fall-back onto in case of an outage. 
A  natural candidate are images from stereo or monocular cameras.  Optical cameras are highly affordable (several orders of magnitude cheaper than LiDAR), operate at a high frame rate, and provide a dense depth map rather than the 64 or 128 sparse rotating laser beams that LiDAR signal is inherently limited to. 

\begin{figure}[t]
  \centerline{\includegraphics[width=\linewidth]{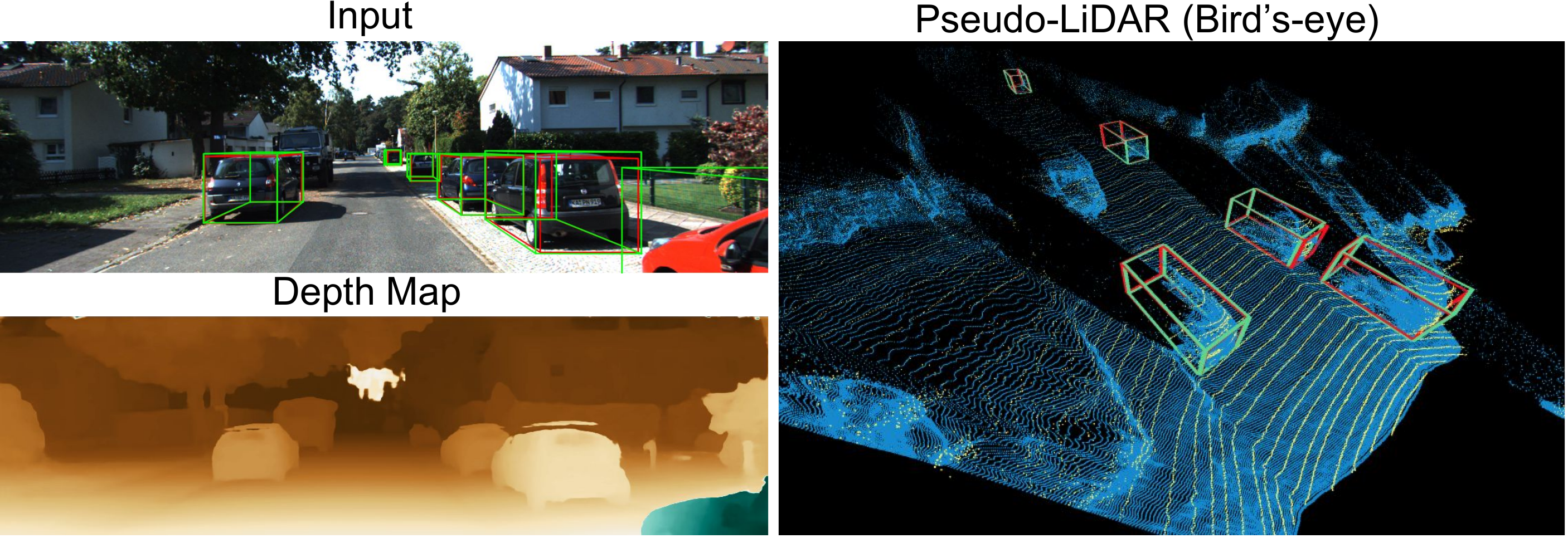}}
  \caption{\textbf{Pseudo-LiDAR signal from visual depth estimation.} Top-left: a KITTI street scene with super-imposed bounding boxes around cars obtained with LiDAR ({\color{red}red}) and pseudo-LiDAR ({\color{green}green}). Bottom-left: estimated disparity map. Right: pseudo-LiDAR ({\color{blue}blue}) vs. LiDAR ({\color{yellow}yellow}) --- the pseudo-LiDAR points align remarkably well with the LiDAR ones. Best viewed in color (zoom in for details.)}
  \label{fig:stere_lidar}
\end{figure}

Several recent publications have explored the use of monocular and stereo depth (disparity) estimation~\cite{godard2017unsupervised, mayer2016large, yamaguchi2014efficient} for 3D object detection~\cite{chen20153d,chen20183d,pham2017robust, xu2018multi}. However, to-date the main successes have been primarily in supplementing LiDAR approaches. For example, one of the leading algorithms~\cite{liang2018deep} on the KITTI benchmark~\cite{geiger2013vision,geiger2012we} uses sensor fusion  to improve the 3D average precision (AP) for cars from 66\% for LiDAR to 73\% with LiDAR and monocular images. 
In contrast, among algorithms that use only images, the state-of-the-art achieves a mere 10\% AP~\cite{xu2018multi}.

One intuitive and popular explanation for such inferior performance is the poor precision of image-based depth estimation. 
In contrast to LiDAR, the error of stereo depth estimation grows quadratically  with depth. However, a visual comparison of the 3D point clouds generated by LiDAR and a state-of-the-art stereo depth estimator~\cite{chang2018pyramid} reveals a high quality match (cf. Fig.~\ref{fig:stere_lidar}) between the two data modalities --- even for faraway objects.

In this paper we provide an alternative explanation with significant performance implications.
We posit that the major cause for the performance gap between stereo and LiDAR is not a discrepancy in depth accuracy, but a poor choice of representations of the 3D information for ConvNet-based 3D object detection systems operating on stereo.
Specifically, the LiDAR signal is commonly represented as 3D point clouds~\cite{qi2018frustum}  or viewed from the top-down ``bird's-eye view'' perspective~\cite{yang2018pixor}, and processed accordingly.
In both cases, the object shapes and sizes are invariant to depth. 
In contrast, image-based depth is densely estimated for each pixel and often represented as additional image channels~\cite{chen20183d,pham2017robust,xu2018multi}, making far-away objects smaller and harder to detect. 
Even worse, pixel neighborhoods in this representation group together points from far-away regions of 3D space.
This makes it hard for convolutional networks relying on 2D convolutions on these channels to reason about and precisely localize objects in 3D.

To evaluate our claim, we introduce a two-step approach for stereo-based 3D object detection. We first convert the estimated depth map from stereo or monocular imagery into a 3D point cloud, which we refer to as \textit{pseudo-LiDAR} as it mimics the LiDAR signal. We then take advantage of existing LiDAR-based 3D object detection pipelines~\cite{ku2018joint,qi2018frustum}, which we  train directly on the pseudo-LiDAR representation. 
By changing the 3D depth representation to pseudo-LiDAR we obtain an unprecedented increase in accuracy of image-based 3D object detection algorithms. 
Specifically, on the KITTI benchmark with IoU (intersection-over-union) at 0.7 for ``moderately hard'' car instances --- the metric used in the official leaderboard --- we achieve a 45.3\% 3D AP on the validation set: almost a 350\% improvement over the previous state-of-the-art image-based approach.
Furthermore, we halve the gap between stereo-based and LiDAR-based systems.

We evaluate multiple combinations of stereo depth estimation and 3D object detection algorithms and arrive at remarkably consistent results. 
This suggests that the gains we observe are because of the \emph{pseudo-LiDAR} representation and are \emph{less dependent} on innovations in 3D object detection architectures or depth estimation techniques.

In sum, the contributions of the paper are two-fold.
First, we show empirically that a major cause for the performance gap between stereo-based and LiDAR-based 3D object detection is not the quality of the estimated depth but its \emph{representation}.
Second, we propose \textit{pseudo-LiDAR} as a new recommended representation of  estimated depth for 3D object detection and show that it leads to state-of-the-art stereo-based 3D object detection, effectively \emph{tripling} prior art.
Our results point towards the possibility of using stereo cameras in self-driving cars --- potentially yielding substantial cost reductions and/or safety improvements. 
\section{Related Work}
\label{related}

\paragraph{LiDAR-based 3D object detection.} Our work is inspired by the recent progress in 3D vision and LiDAR-based 3D object detection. 
Many recent techniques use the fact that LiDAR is naturally represented as 3D point clouds. For example, frustum PointNet~\cite{qi2018frustum} applies PointNet~\cite{qi2017pointnet} to each frustum proposal from a 2D object detection network.
MV3D~\cite{chen2017multi} projects LiDAR points into both bird-eye view (BEV) and frontal view to obtain multi-view features. VoxelNet~\cite{zhou2018voxelnet} encodes 3D points into voxels and extracts features by 3D convolutions. UberATG-ContFuse~\cite{liang2018deep}, one of the leading algorithms on the KITTI benchmark~\cite{geiger2012we}, performs continuous convolutions~\cite{wang2018deep} to fuse visual and BEV LiDAR features. All these algorithms assume that the precise 3D point coordinates are given. The main challenge there is thus on predicting point labels or drawing bounding boxes in 3D to locate objects. 

\paragraph{Stereo- and monocular-based depth estimation.}
A key ingredient for image-based 3D object detection methods is a reliable depth estimation approach to replace LiDAR. These can be obtained through monocular~\cite{fu2018deep,godard2017unsupervised} or stereo vision~\cite{chang2018pyramid,mayer2016large}. 
The accuracy of these systems has increased dramatically since early work on monocular depth estimation~\cite{eigen2014depth, KevinKarsch, saxena2009make3d}. Recent algorithms like DORN~\cite{fu2018deep} combine multi-scale features with ordinal regression to predict pixel depth with remarkably low errors. For stereo vision, PSMNet~\cite{chang2018pyramid} applies Siamese networks for disparity estimation, followed by 3D convolutions for refinement, resulting in an outlier rate less than $2\%$. 
Recent work has made these methods mode efficient~\cite{wang2018anytime}, enabling accurate disparity estimation to run at 30 FPS on mobile devices.

\paragraph{Image-based 3D object detection.}
The rapid progress on stereo and monocular depth estimation suggests that they could be used as a substitute for LiDAR in image-based 3D object detection algorithms.
Existing algorithms of this flavor are largely built upon 2D object detection~\cite{ren2015faster}, imposing extra geometric constraints~\cite{chabot2017deep,chen2016monocular,mousavian20173d,xiang2017subcategory} to create 3D proposals. \cite{chen20153d,chen20183d,pham2017robust,xu2018multi} apply stereo-based depth estimation to obtain the true 3D coordinates of each pixel. These 3D coordinates are either entered as additional input channels into a 2D detection pipeline, or used to extract hand-crafted features. 
Although these methods have made remarkable progress, the state-of-the-art for 3D object detection performance lags behind LiDAR-based methods.
As we discuss in Section~\ref{approach}, this might be because of the depth representation used by these methods.
\section{Approach}
\label{approach}

\begin{figure*}
	\centering
	\includegraphics[width=\linewidth]{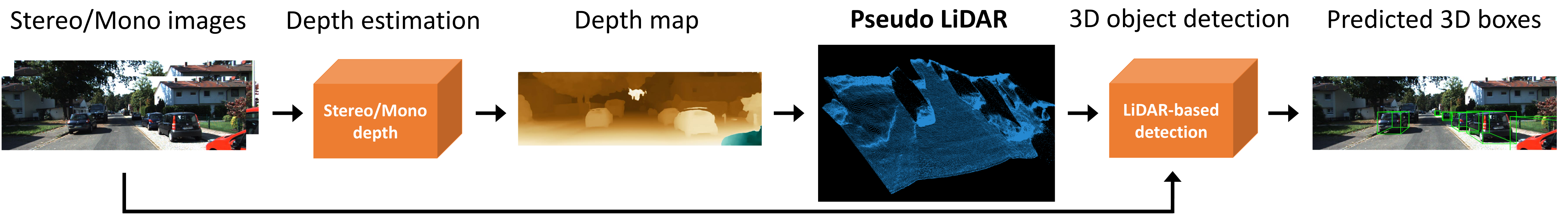}
	\caption{\textbf{The proposed pipeline for image-based 3D object detection.} Given stereo or monocular images, we first predict the depth map, followed by back-projecting it into a 3D point cloud in the LiDAR coordinate system. We refer to this representation as \emph{pseudo-LiDAR}, and process it exactly like LiDAR --- any LiDAR-based detection algorithms can be applied.}
	\label{fig:pipeline}
\end{figure*}

Despite the many advantages of image-based 3D object recognition,  there remains a glaring gap between the state-of-the-art detection rates of image and LiDAR-based approaches (see Table \ref{tbMain} in Section~\ref{exp_result}).  It is tempting to attribute this gap to the obvious physical differences and its implications between LiDAR and camera technology.  For example, the error of stereo-based 3D depth estimation grows quadratically with the depth of an object, whereas for Time-of-Flight (ToF) approaches, such as LiDAR, this relationship is approximately linear.

Although some of these physical differences do likely contribute to the accuracy gap, in this paper we claim that a large portion of the discrepancy can be explained by the data representation rather than its quality or underlying physical properties associated with data collection.

In fact, recent algorithms for stereo depth estimation can generate surprisingly accurate depth maps~\cite{chang2018pyramid} (see figure~\ref{fig:stere_lidar}).  Our approach to ``close the gap'' is therefore to carefully remove the differences between the two data modalities and align the two recognition pipelines as much as possible.
To this end, we propose a two-step approach by first estimating the dense pixel depth from stereo (or even monocular) imagery and then back-projecting pixels into a 3D point cloud. By viewing this representation as \emph{pseudo-LiDAR} signal, we can then apply \emph{any} existing LiDAR-based 3D object detection algorithm. Fig.~\ref{fig:pipeline} depicts our pipeline.

\paragraph{Depth estimation.}
Our approach is agnostic to different depth estimation algorithms.
We primarily work with stereo disparity estimation algorithms~\cite{chang2018pyramid,mayer2016large}, although our approach can easily use monocular depth estimation methods.

A stereo disparity estimation algorithm takes a pair of left-right images $I_l$ and $I_r$ as input, captured from a pair of cameras with a horizontal offset (i.e., baseline) $b$, and outputs a disparity map $Y$ of the same size as either one of the two input images. Without loss of generality, we assume the depth estimation algorithm treats the left image, $I_l$, as reference and records in $Y$ the horizontal disparity to $I_r$ for each pixel. Together with the horizontal focal length $f_U$ of the left camera, we can derive the depth map $D$ via the following transform,
\begin{align}
D(u, v) = \frac{f_U\times b}{Y(u, v)}. \label{eq_disp_depth}
\end{align}
 
\paragraph{Pseudo-LiDAR generation.}
Instead of incorporating the depth $D$ as multiple additional channels to the RGB images, as is typically done~\cite{xu2018multi}, we can derive the 3D location $(x, y, z)$ of each pixel $(u, v)$, in the left camera's coordinate system, as follows,
\begin{align}
& \text{(depth)} \hspace{20pt} z = D(u, v)\\
& \text{(width)} \hspace{20pt} x = \frac{(u - c_U)\times z}{f_U}\\
& \text{(height)} \hspace{20pt} y = \frac{(v - c_V)\times z}{f_V},
\end{align}
where $(c_U, c_V)$ is the pixel location corresponding to the camera center and $f_V$ is the vertical focal length.

By back-projecting all the pixels into 3D coordinates, we arrive at a 3D point cloud $\{(x^{(n)}, y^{(n)}, z^{(n)})\}_{n=1}^N$, where $N$ is the pixel count. Such a point cloud can be transformed into any cyclopean coordinate frame given a reference viewpoint and viewing direction. We refer to the resulting point cloud as \emph{pseudo-LiDAR} signal.

\paragraph{LiDAR vs. pseudo-LiDAR.}
 In order to be maximally compatible with existing LiDAR detection pipelines we apply a few additional post-processing steps on the pseudo-LiDAR data. Since real LiDAR signals only reside in a certain range of heights, we disregard pseudo-LiDAR points beyond that range. For instance, on the KITTI benchmark, following~\cite{yang2018pixor}, we remove all points higher than 1m above the fictitious LiDAR source (located on top of the autonomous vehicle). As most objects of interest (e.g., cars and pedestrians)  do not exceed this height range there is little information loss.
  In addition to depth, LiDAR also returns the reflectance of any measured pixel (within [0,1]).  As we have no such information,  we simply set the reflectance to 1.0 for every pseudo-LiDAR points.

Fig~\ref{fig:stere_lidar} depicts the ground-truth LiDAR and the pseudo-LiDAR points for the same scene from the KITTI dataset~\cite{geiger2013vision, geiger2012we}. The depth estimate was obtained with the pyramid stereo matching network (PSMNet)~\cite{chang2018pyramid}.  Surprisingly, the pseudo-LiDAR points ({\color{blue}blue}) align remarkably well to true LiDAR points ({\color{yellow}yellow}), in contrast to the common belief that low precision image-based depth is the main cause of inferior 3D object detection. We note that a LiDAR can capture $>100,000$ points for a scene, which is of the same order as the pixel count. Nevertheless, LiDAR points are distributed along a few (typically 64 or 128) horizontal beams, only sparsely occupying the 3D space.

\paragraph{3D object detection.} With the estimated pseudo-LiDAR points, we can apply \emph{any} existing LiDAR-based 3D object detectors for autonomous driving. In this work, we consider those based on multimodal information (i.e., monocular images + LiDAR), as it is only natural to incorporate the original visual information together with the pseudo-LiDAR data. Specifically, we experiment on AVOD~\cite{ku2018joint} and frustum PointNet~\cite{qi2018frustum}, the two top ranked algorithms with open-sourced code on the KITTI benchmark.
In general, we distinguish between two different setups:
\begin{itemize}
\item[a)] In the first setup we treat the pseudo-LiDAR information as a \emph{3D point cloud}. Here,  we use frustum PointNet~\cite{qi2018frustum}, which projects 2D object detections~\cite{lin2017feature} into a frustum in 3D, and then applies PointNet~\cite{qi2017pointnet} to extract point-set features at each 3D frustum.

\item[b)] In the second setup we view the pseudo-LiDAR information from a \emph{Bird's Eye View (BEV)}.  In particular, the 3D information is converted into a 2D image from the top-down view: width and depth become the spatial dimensions, and height is recorded in the channels.   AVOD connects visual features and BEV LiDAR features to 3D box proposals and then fuses both to perform box classification and regression.
\end{itemize}

\paragraph{Data representation matters.}
 Although pseudo-LiDAR conveys the same information as a depth map, we claim that it is much better suited for  3D object detection pipelines that are based on deep convolutional networks.
 To see this, consider the core module of the convolutional network: 2D convolutions.
 A convolutional network operating on images or depth maps performs a sequence of 2D convolutions on the image/depth map.
 Although the filters of the convolution can be learned, the central assumption is two-fold: (a) local neighborhoods in the image have meaning, and the network should look at local patches, and (b)  all neighborhoods can be operated upon in an identical manner.

 These are but imperfect assumptions.
 First, local patches on 2D images are only coherent physically if they are entirely contained in a single object.
 If they straddle object boundaries, then two pixels can be co-located next to each other in the depth map, yet can be very far away in 3D space.
 Second, objects that occur at multiple depths project to \emph{different scales} in the depth map.
 A similarly sized patch might capture just a side-view mirror of a nearby car or the entire body of a far-away car.
 Existing 2D object detection approaches struggle with this breakdown of assumptions and have to design novel techniques such as feature pyramids~\cite{lin2017feature} to deal with this challenge.
 
 \begin{figure}
 	\centering
 	\includegraphics[width=\columnwidth]{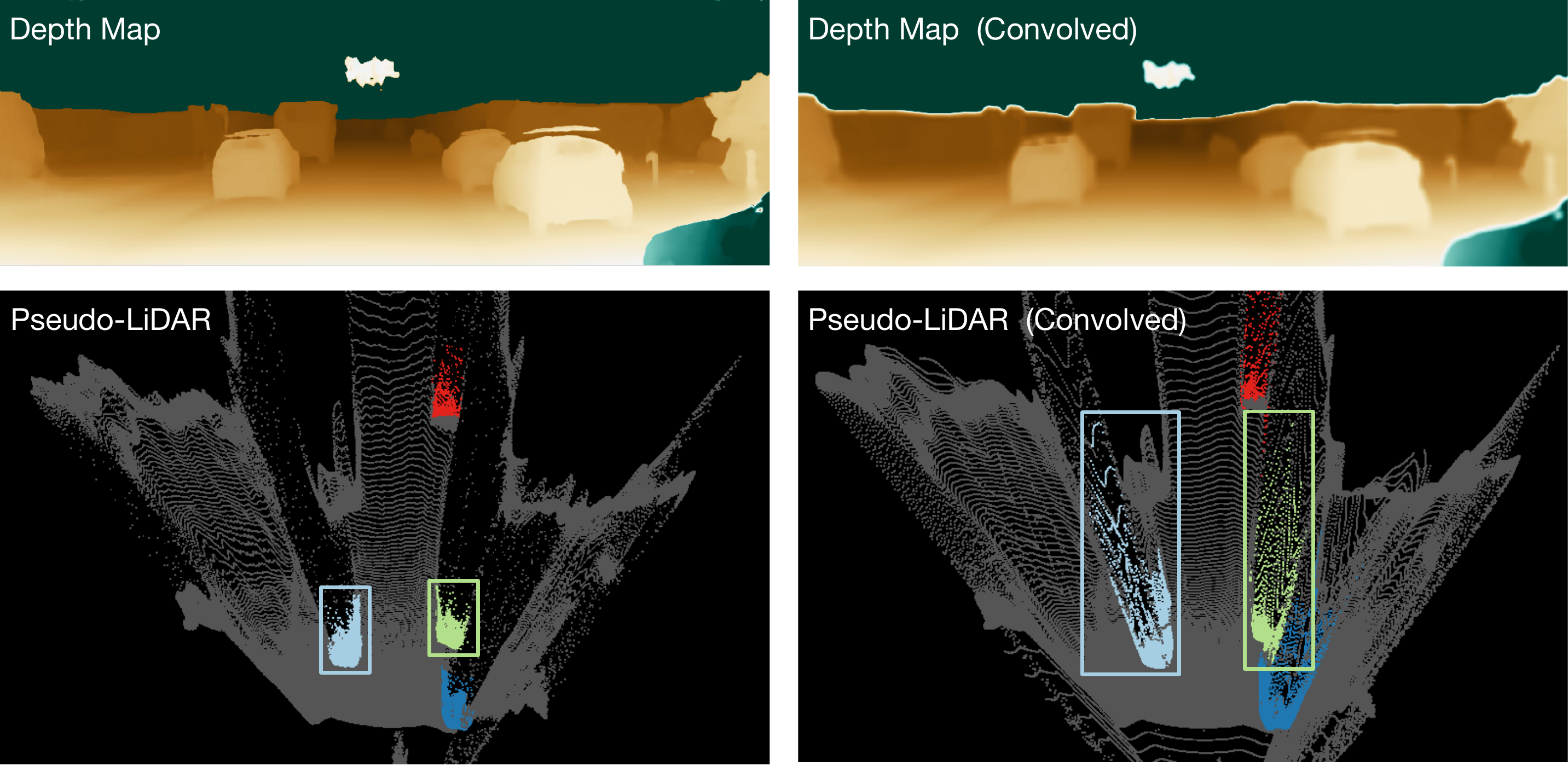}
 	\caption{We apply a single 2D convolution with a uniform kernel to the frontal view depth map (top-left). The resulting depth map (top-right), after back-projected into pseudo-LiDAR and displayed from the bird's-eye view (bottom-right), reveals a large depth distortion in comparison to the original pseudo-LiDAR representation (bottom-left), especially for far-away objects. We mark points of each car instance by a color. The boxes are super-imposed and contain all points of the green and cyan cars respectively. }
 	\label{fig:depth_conv}
 \end{figure}

 In contrast, 3D convolutions on point clouds or 2D convolutions in the bird's-eye view slices operate on pixels that are \emph{physically} close together (although the latter do pull together pixels from different heights,  the physics of the world implies that pixels at different heights at a particular spatial location usually do belong to the same object).
 In addition, both far-away objects and nearby objects are treated exactly the same way.
 These operations are thus inherently more physically meaningful and hence should lead to better learning and more accurate models.

To illustrate this point further, in Fig.~\ref{fig:depth_conv} we conduct a simple experiment.  In the left column, we show the original depth-map and the pseudo-LiDAR representation of an image scene.  The four cars in the scene are highlighted in color.
We then perform a single $11\times 11$ convolution with a box filter on the depth-map (top right), which matches the receptive field of 5 layers of $3 \times 3$ convolutions. We then convert the resulting (blurred) depth-map  into a pseudo-LiDAR representation (bottom right).  From the figure, it becomes evident that this new pseudo-LiDAR representation suffers substantially from the effects of the blurring. The cars are stretched out far beyond their actual physical proportions making it essentially impossible to locate them precisely.  For better visualization, we added rectangles that contain all the points of the green and cyan cars.  After the convolution, both bounding boxes capture highly erroneous areas. Of course, the 2D convolutional network will learn to use more intelligent filters than box filters, but this example goes to show how some operations the convolutional network might perform could border on the absurd.

\section{Experiments}
\label{exp}
We evaluate   3D-object detection with and without pseudo-LiDAR across different settings with varying approaches for depth estimation and object detection. Throughout, we will highlight results obtained with \emph{pseudo-LiDAR} in {\color{blue} blue} and those with actual LiDAR in {\color{gray} gray}.

\subsection{Setup}
\label{exp_setup}

\paragraph{Dataset.}
We evaluate our approach on the KITTI object detection benchmark~\cite{geiger2013vision,geiger2012we}, which contains 7,481 images for training and 7,518 images for testing. We follow the same training and validation splits as suggested by Chen et al.~\cite{chen20153d}, containing 3,712 and 3,769 images respectively. For each image, KITTI provides the corresponding Velodyne LiDAR point cloud, right image for stereo information, and camera calibration matrices.

\paragraph{Metric.}
We focus on 3D and bird's-eye-view (BEV)\footnote{The BEV detection task is also called 3D localization.} object detection and report the results on the \emph{validation set}. Specifically, we focus on the ``car'' category, following~\cite{chen2017multi, xu2018pointfusion}. We follow the benchmark and prior work and report average precision (AP) with the IoU thresholds at 0.5 and 0.7. We denote AP for the 3D and BEV tasks by \AP and \APBEV, respectively. Note that the benchmark divides each category into three cases --- easy, moderate, and hard --- according to the bounding box height and occlusion/truncation level. In general, the easy case corresponds to cars within 30 meters of the ego-car distance~\cite{yang2018pixor}.

\paragraph{Baselines.}
We compare to \Mono~\cite{chen2016monocular}, \DOP~\cite{chen20153d}, and \MLF~\cite{xu2018multi}. The first is monocular and the second is stereo-based. \MLF~\cite{xu2018multi} reports results with both monocular~\cite{godard2017unsupervised} and stereo disparity~\cite{mayer2016large}, which we denote as \MLFmono and \MLFstereo, respectively.

\subsection{Details of our approach}
\label{exp_approach}
\paragraph{Stereo disparity estimation.}
We apply \PSMNet~\cite{chang2018pyramid}, \DispNet~\cite{mayer2016large}, and \SPSstereo~\cite{yamaguchi2014efficient} to estimate dense disparity. The first two approaches are learning-based and we use the released models, which are pre-trained on the Scene Flow dataset~\cite{mayer2016large}, with over 30,000 pairs of synthetic images and dense disparity maps, and fine-tuned on the 200 training pairs of KITTI stereo 2015 benchmark~\cite{geiger2012we,menze2015object}. We note that, \MLFstereo~\cite{xu2018multi} also uses the released \DispNet model. The third approach, \SPSstereo~\cite{yamaguchi2014efficient}, is non-learning-based and has been used in~\cite{chen20153d,chen20183d,pham2017robust}.

\DispNet has two versions, without and with correlations layers. We test both and denote them as \DispNetS and \DispNetC, respectively.

While performing these experiments, we found that the 200 training images of KITTI stereo 2015 overlap with the validation images of KITTI object detection.
That is, the released \PSMNet and \DispNet models actually used some validation images of detection.
We therefore train a version of \PSMNet using Scene Flow followed by finetuning on the 3,712 training images of detection, instead of the 200 KITTI stereo images.
We obtain \emph{pseudo disparity} ground truth by projecting the corresponding LiDAR points into the 2D image space. We denote this version \PSMNetpd. Details are included in the Supplementary Material.

The results with \PSMNetpd in Table~\ref{tbDisp} (fined-tuned on 3,712 training data) are in fact better than \PSMNet (fine-tuned on KITTI stereo 2015). We attribute the improved accuracy of \PSMNetpd  on the fact that it is trained on a larger training set. Nevertheless, future work on 3D object detection using stereo must be aware of this overlap.

\paragraph{Monocular depth estimation.} We use the state-of-the-art monocular depth estimator \DORN~\cite{fu2018deep}, which is trained by the authors on 23,488 KITTI images. We note that some of these images may overlap with our validation data for detection. Nevertheless,  we decided to still include these results and believe they could serve as an upper bound for monocular-based 3D object detection. Future work, however, must be aware of this overlap.

\paragraph{Pseudo-LiDAR generation.}
We back-project the estimated depth map into 3D points in the Velodyne LiDAR's coordinate system using the provided calibration matrices. We disregard points with heights larger than 1 in the system.

\paragraph{3D Object detection.} We consider two algorithms: Frustum PointNet (\Frustum)~\cite{qi2018frustum} and \AVOD~\cite{ku2018joint}. More specifically, we apply \Frustum-v1 and \AVOD-FPN. Both of them use information from LiDAR and monocular images. We train both models on the 3,712 training data from scratch by replacing the LiDAR points with pseudo-LiDAR data generated from stereo disparity estimation. We use the hyper-parameters provided in the released code. 

We note that \AVOD takes image-specific ground planes as inputs. The authors provide ground-truth planes for training and validation images, but do not provide the procedure to obtain them (for novel images). We therefore fit the ground plane parameters with a straight-forward application of RANSAC~\cite{fischler1981random} to our pseudo-LiDAR points that fall into a certain range of road height, during evaluation. Details are included in the Supplementary Material.
\begin{table*}[t]
	\centering
	\caption{\small 3D object detection results on the KITTI validation set. We report \APBEV ~/ \AP (in \%) of the \textbf{car} category, corresponding to average precision of the bird's-eye view and 3D object box detection. Mono stands for monocular. Our methods with \emph{pseudo-LiDAR} estimated by \PSMNetpd~\cite{chang2018pyramid} (stereo) or \DORN~\cite{fu2018deep} (monocular) are in {\color{blue} blue}. Methods with LiDAR are in {\color{gray} gray}. Best viewed in color.} \label{tbMain}
	\begin{tabular}{=l|+c|+c|+c|+c|+c|+c|+c}
		&  & \multicolumn{3}{c|}{IoU = 0.5} & \multicolumn{3}{c}{IoU = 0.7} \\ \cline{3-8}
		\multicolumn{1}{c|}{Detection algorithm} & Input signal & Easy & Moderate & Hard & Easy & Moderate & Hard \\ \hline
		\Mono~\cite{chen2016monocular} & Mono & 30.5 / 25.2 & 22.4 / 18.2 & 19.2 / 15.5  & 5.2 / 2.5 & 5.2 / 2.3  & 4.1 / 2.3  \\
		\MLFmono~\cite{xu2018multi} & Mono & 55.0 / 47.9 & 36.7 / 29.5 & 31.3 / 26.4 & 22.0 / 10.5 & 13.6 / 5.7 & 11.6 / 5.4\\
		\rowstyle{\color{blue}}
		\AVODC & Mono & 61.2 / 57.0 & 45.4 / 42.8 & 38.3 / 36.3 & 33.7 / 19.5 & 24.6 / 17.2 & 20.1 / 16.2\\
		\rowstyle{\color{blue}}
		\Frustum & Mono & 70.8 / 66.3 & 49.4 / 42.3 & 42.7 / 38.5 & 40.6 / 28.2 & 26.3 / 18.5 & 22.9 / 16.4\\ \hline
		\DOP~\cite{chen20153d} & Stereo & 55.0 / 46.0 & 41.3 / 34.6 & 34.6 / 30.1 & 12.6 / 6.6 & 9.5 / 5.1 & 7.6 / 4.1 \\ 
		\MLFstereo~\cite{xu2018multi} & Stereo & - & 53.7 / 47.4 & - & - & 19.5 / 9.8 & - \\
		\rowstyle{\color{blue}}
		\AVODC & Stereo & 89.0 / 88.5 & 77.5 / 76.4 & 68.7 / 61.2&  74.9 / 61.9 &  56.8 / 45.3 & 49.0 / 39.0 \\
		\rowstyle{\color{blue}}
		\rowstyle{\color{blue}}
		\Frustum & Stereo & 89.8 / 89.5 & 77.6 / 75.5 &  68.2 / 66.3 &  72.8 / 59.4& 51.8 / 39.8 & 44.0 / 33.5 \\ \hline
		\rowstyle{\color{gray}}
		\AVODGT~\cite{ku2018joint} & LiDAR + Mono & 90.5 / 90.5 & 89.4 / 89.2 & 88.5 / 88.2 & 89.4 / 82.8 & 86.5 / 73.5 & 79.3 / 67.1 \\
		\rowstyle{\color{gray}}
		\Frustum~\cite{qi2018frustum} & LiDAR + Mono & 96.2 / 96.1 & 89.7 /  89.3 & 86.8 / 86.2 &  88.1 / 82.6 & 82.2 / 68.8 & 74.0 / 62.0 \\
		\hline
	\end{tabular}
\end{table*}

\subsection{Experimental results}
\label{exp_result}

We summarize the main results in Table~\ref{tbMain}. We organize methods according to the input signals for performing detection. Our stereo approaches based on pseudo-LiDAR significantly outperform all image-based alternatives by a large margin. At IoU = 0.7 (moderate) --- the metric used to rank algorithms on the KITTI leaderboard --- we achieve \emph{double} the performance of the previous state of the art. We also observe that pseudo-LiDAR is applicable and highly beneficial to two 3D object detection algorithms with very different architectures, suggesting its wide compatibility.

One interesting comparison is between approaches using pseudo-LiDAR with monocular depth (\DORN) and stereo depth (\PSMNetpd). While \DORN has been trained with almost ten times more images than \PSMNetpd (and some of them overlap with the validation data), the results with \PSMNetpd dominate. This suggests that stereo-based detection is a promising direction to move in, especially considering the increasing affordability of stereo cameras.

In the following section, we discuss key observations and conduct a series of experiments to analyze the performance gain through pseudo-LiDAR with stereo disparity.

\begin{table}
	\centering
	\tabcolsep 3pt
	\caption{\small Comparison between frontal and \emph{pseudo-LiDAR} representations. \AVOD projects the pseudo-LiDAR representation into the bird-eye's view (BEV). We report \APBEV ~/ \AP (in \%) of the \textbf{moderate car} category at IoU = 0.7. The best result of each column is in bold font. The results indicate strongly  that the data representation is the key  contributor to the accuracy gap. } \label{tbAVOD}
	\begin{tabular}{l|l|c|c}
		\multicolumn{1}{c|}{Detection} & \multicolumn{1}{c|}{Disparity} & Representation & \APBEV~/ \AP \\ \hline
		\MLF~\cite{xu2018multi} & \DispNet & Frontal & 19.5 / 9.8 \\
		\color{blue}\AVODGT &\color{blue} \DispNetS &\color{blue} Pseudo-LiDAR & \color{blue}36.3 / 27.0 \\
		\color{blue}\AVODGT &\color{blue} \DispNetC &\color{blue} Pseudo-LiDAR & \color{blue}36.5 / 26.2 \\ \hline
		\AVODGT & \PSMNetpd & Frontal & 11.9 / 6.6 \\
		\color{blue}\AVODGT &\color{blue} \PSMNetpd &\color{blue} Pseudo-LiDAR & \color{blue}\textbf{56.8} / \textbf{45.3} \\
		\hline
	\end{tabular}
\end{table} 

\paragraph{Impact of data representation.}
When comparing our results using \DispNetS or \DispNetC to \MLFstereo~\cite{xu2018multi} (which also uses \DispNet as the underlying stereo engine), we observe a large performance gap (see Table.~\ref{tbAVOD}). Specifically, at IoU$ = 0.7$,  we outperform \MLFstereo by at least 16\% on \APBEV and 16\% on \AP. The later is equivalent to a 160\% relative improvement. 
We attribute this improvement to the way in which we represent the resulting depth information. We note that both our approach and \MLFstereo~\cite{xu2018multi} first back-project pixel depths into 3D point coordinates. \MLFstereo construes the 3D coordinates of each pixel as additional feature maps in the frontal view. These maps are then concatenated with RGB channels as the input to a modified 2D object detection pipeline based on Faster-RCNN~\cite{ren2015faster}. As we point out earlier, this has two problems. Firstly, distant objects become smaller, and detecting small objects is a known hard problem~\cite{lin2017feature}.
Secondly, while performing local computations like convolutions or ROI pooling along height and width of an image makes sense to 2D object detection, it will operate on 2D pixel neighborhoods with pixels that are far apart in 3D, making the precise localization of 3D objects much harder (cf.~Fig.~\ref{fig:depth_conv}).

By contrast, our approach treats these coordinates as pseudo-LiDAR signals and applies PointNet~\cite{qi2017pointnet} (in \Frustum) or use a convolutional network on the BEV projection (in \AVOD).
This introduces invariance to depth, since far-away objects are no longer smaller.
Furthermore, convolutions and pooling operations in these representations put together points that are physically nearby.
 
To further control for other differences between \MLFstereo and our method we ablate our approach to use the same frontal depth representation used by \MLFstereo. 
\AVOD fuses information of the frontal images with BEV LiDAR features. We modify the algorithm, following \cite{chen20183d,xu2018multi}, to generate five frontal-view feature maps, including 3D pixel locations, disparity, and Euclidean distance to the camera. We concatenate them with the RGB channels while disregarding the BEV branch in \AVOD, making it fully dependent on the frontal-view branch. (We make no additional architecture changes.) The results in Table~\ref{tbAVOD} reveal a staggering gap between frontal and pseudo-LiDAR results. We found that the frontal approach  struggles with inferring object depth, even when the five extra maps have provided sufficient 3D information.
Again, this might be because 2d convolutions put together pixels from far away depths, making accurate localization difficult. This experiment suggests that the chief source of the accuracy improvement is indeed the \emph{pseudo-LiDAR} representation.

\begin{table}
	\centering
	\caption{\small Comparison of different combinations of stereo disparity and 3D object detection algorithms, using \emph{pseudo-LiDAR}. We report \APBEV ~/ \AP (in \%) of the \textbf{moderate car} category at IoU = 0.7. The best result of each column is in bold font.} \label{tbDisp}
	\begin{tabular}{l|c|c}
		\multicolumn{1}{c|}{}  & \multicolumn{2}{c}{Detection algorithm} \\ \cline{2-3}
		\multicolumn{1}{c|}{Disparity} & \AVODGT & \Frustum \\ \hline
		\color{blue}\DispNetS &\color{blue} 36.3 / 27.0  &\color{blue} 31.9 / 23.5 \\
		\color{blue}\DispNetC  &\color{blue} 36.5 / 26.2 &\color{blue} 37.4 / 29.2\\
		\color{blue}\PSMNet  &\color{blue} 39.2 / 27.4 &\color{blue} 33.7 / 26.7 \\ \hline
		\color{blue}\PSMNetpd  &\color{blue} \textbf{56.8} / \textbf{45.3} &\color{blue} \textbf{51.8} / \textbf{39.8} \\
		\hline
	\end{tabular}
\end{table}

\paragraph{Impact of stereo disparity estimation accuracy.}

We compare \PSMNet~\cite{chang2018pyramid} and \DispNet~\cite{mayer2016large} on pseudo-LiDAR-based detection accuracies. On the leaderboard of KITTI stereo 2015, \PSMNet achieves 1.86\% disparity error, which far outperforms the error of 4.32\% by \DispNetC.

As shown in Table~\ref{tbDisp}, the accuracy of disparity estimation does not necessarily correlate with the accuracy of object detection. \Frustum with \DispNetC even outperforms \Frustum with \PSMNet.
This is likely due to two reasons. First, the disparity accuracy may not reflect the depth accuracy: the same disparity error (on a pixel) can lead to drastically different depth errors dependent on the pixel's true depth, according to Eq.~(\ref{eq_disp_depth}). Second, different detection algorithms process the 3D points differently: \AVOD quantizes points into voxels, while \Frustum directly processes them and may be vulnerable to noise. 

By far the most accurate detection results are obtained by \PSMNetpd, which we trained from scratch on our own KITTI training set.  These results seem to suggest that significant further improvements may be possible through  end-to-end training  of the whole pipeline. 

We provide results using \SPSstereo~\cite{yamaguchi2014efficient} and further analysis on depth estimation in the Supplementary Material.

\paragraph{Comparison to LiDAR information.}
Our approach significantly improves stereo-based detection accuracies. 
A key remaining question is, how close the pseudo-LiDAR detection results are to those based on real LiDAR signal. 
In Table~\ref{tbMain}, we further compare to \AVOD and \Frustum when actual LiDAR signal is available. For fair comparison, we retrain both models. 
For the easy cases with IoU $=0.5$, our stereo-based approach performs very well, only slightly worse than the corresponding LiDAR-based version. However, as the instances become harder (e.g., for cars that are far away), the performance gaps resurfaces --- although  not nearly as pronounced as without pseudo-LiDAR. 
We also see a larger gap when moving to IoU $=0.7$.
These results are not surprising, since stereo algorithms are known to have larger depth errors for far-away objects, and a stricter metric requires higher depth precision. Both observations emphasize the need for accurate depth estimation, especially for far-away distances, to bridge the gap further.  A key limitation of our results may be the low resolution of the 0.4 MegaPixel images, which  cause far away objects to only consist of a few pixels. 

\begin{table}
\centering
\tabcolsep 5pt
\caption{\small 3D object detection on the \textbf{pedestrian} and \textbf{cyclist} categories on the validation set. We report \APBEV~/ \AP at IoU = 0.5 (the standard metric) and compare \Frustum with \emph{pseudo-LiDAR} estimated by \PSMNetpd (in {\color{blue}blue}) and LiDAR (in {\color{gray} gray}). } \label{tbPedestrian}
\begin{tabular}{=c|+c|+c|+c}
Input signal & Easy & Moderate & Hard \\ \hline
\multicolumn{4}{c}{Pedestrian} \\ \hline
\rowstyle{\color{blue}}
Stereo & 41.3 / 33.8 & 34.9 / 27.4 & 30.1 / 24.0\\
\rowstyle{\color{gray}}
LiDAR + Mono & 69.7 / 64.7 & 60.6 / 56.5 & 53.4 / 49.9 \\ \hline
\multicolumn{4}{c}{Cyclist} \\ \hline
\rowstyle{\color{blue}}
Stereo & 47.6 / 41.3 & 29.9 / 25.2 & 27.0 / 24.9\\
\rowstyle{\color{gray}}
LiDAR + Mono & 70.3 / 66.6 & 55.0 / 50.9 & 52.0 / 46.6\\
\hline
\end{tabular}
\end{table}

\begin{figure*}[t]
	\centerline{\includegraphics[width=\linewidth]{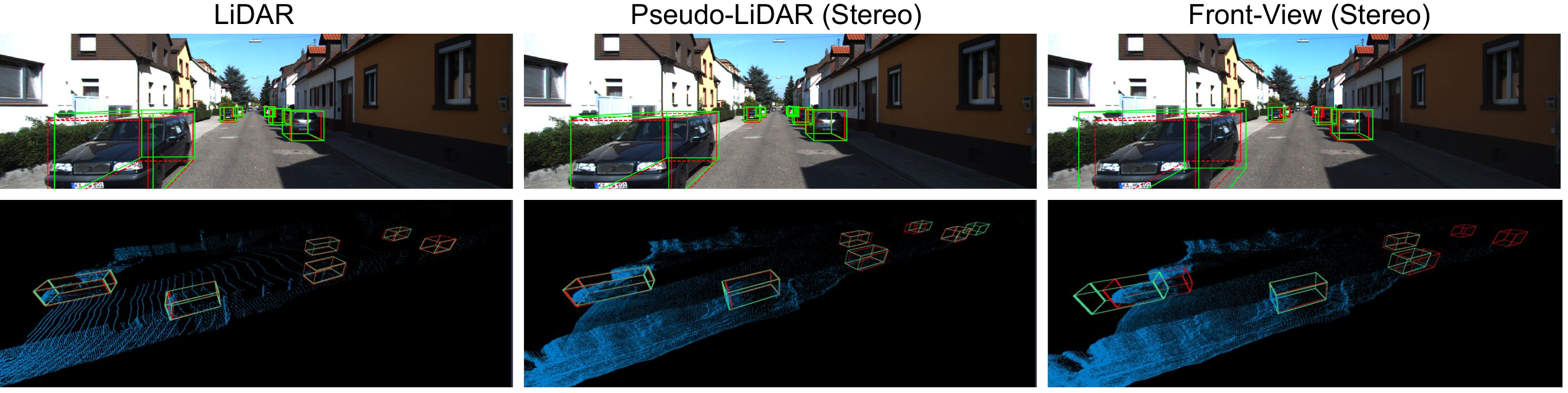}}
	\caption{\textbf{Qualitative comparison.} We compare \AVOD with LiDAR, pseudo-LiDAR, and frontal-view (stereo). Ground-truth boxes are in {\color{red}{red}}, predicted boxes in {\color{green}{green}}; the observer in the pseudo-LiDAR plots (bottom row) is on the very left side looking to the right.  The frontal-view approach (\emph{right}) even  miscalculates the depths of nearby objects and misses far-away objects entirely. Best viewed in color.}
	\label{fig:qualitative}
\end{figure*}

\paragraph{Pedestrian and cyclist detection.}
We also present results on 3D pedestrian and cyclist detection.
These are much more challenging tasks than car detection due to the small sizes of the objects, even given LiDAR signals. At an IoU threshold of 0.5, both \APBEV and \AP of pedestrians and cyclists are much lower than that of cars at IoU 0.7~\cite{qi2018frustum}. We also notice that none of the prior work on image-based methods report  results in this category.

Table~\ref{tbPedestrian} shows our results with \Frustum and compares to those with LiDAR, on the validation set. Compared to the car category (cf.~Table~\ref{tbMain}), the performance gap is significant. We also observe a similar trend that the gap becomes larger when moving to the hard cases. Nevertheless, our approach has set a solid starting point for image-based pedestrian and cyclist detection for future work.

\subsection{Results on the test set}
We report our results on the car category on the test set in Table~\ref{tbTest}. We see a similar gap between pseudo-LiDAR and LiDAR as on the validation set, suggesting that our approach does not simply over-fit to the ``validation data.'' \emph{We also note that, at the time we submit the paper, we are at the first place among all the image-based algorithms on the KITTI leaderboard.} Details and results on the pedestrian and cyclist categories are in the Supplementary Material.

\begin{table}
	\centering
	\tabcolsep 5pt
	\caption{\small 3D object detection results on the \textbf{car} category on the \emph{test} set. We compare \emph{pseudo-LiDAR} with \PSMNetpd (in {\color{blue}blue}) and LiDAR (in {\color{gray} gray}). We report \APBEV~/ \AP at IoU = 0.7.
	$\dagger$: Results on the KITTI leaderboard.} \label{tbTest}
	\begin{tabular}{=c|+c|+c|+c}
		Input signal & Easy & Moderate & Hard \\ \hline
		\multicolumn{4}{c}{\AVOD} \\ \hline
		\rowstyle{\color{blue}}
		Stereo &  66.8 / 55.4 &  47.2 / 37.2 & 40.3 / 31.4\\
		\rowstyle{\color{gray}}
		$\dagger$LiDAR + Mono  &  88.5 / 81.9 &  83.8 / 71.9 & 77.9 / 66.4\\
		\hline
		\multicolumn{4}{c}{\Frustum} \\ \hline
		\rowstyle{\color{blue}}
		Stereo &  55.0 / 39.7 &  38.7 / 26.7 & 32.9 / 22.3\\
		\rowstyle{\color{gray}}
		$\dagger$LiDAR + Mono  &  88.7 / 81.2 &  84.0 / 70.4 & 75.3 / 62.2\\
		\hline
	\end{tabular}
\end{table}

\subsection{Visualization} We further visualize the prediction results on validation images in Fig.~\ref{fig:qualitative}. We compare LiDAR (left), stereo pseudo-LiDAR (middle),
and frontal stereo (right). We used \PSMNetpd to obtain the stereo depth maps. 
LiDAR and pseudo-LiDAR lead to highly accurate predictions, especially for the nearby objects.  However, pseudo-LiDAR fails to detect far-away objects precisely due to inaccurate depth estimates.
On the other hand, the frontal-view-based approach makes extremely inaccurate predictions, even for nearby objects. This corroborates the quantitative results we observed in Table~\ref{tbAVOD}. We provide additional qualitative results and failure cases in the Supplementary Material.

\section{Discussion and Conclusion}
\label{disc}

Sometimes, it is the simple discoveries that make the biggest differences. In this paper we have shown that  a key component to closing the gap between image- and LiDAR-based 3D object detection  may be simply the representation of the 3D information.  It may be fair to consider these results as the correction of a systemic inefficiency rather than  a novel algorithm --- however, that does not diminish its importance.  Our findings are consistent with  our understanding of convolutional neural networks and substantiated through empirical results. In fact, the improvements we obtain from this correction are unprecedentedly high and affect all methods alike.  With this quantum leap it is  plausible that image-based 3D object detection for autonomous vehicle will become a reality in the near future.  The implications of such a prospect are enormous. Currently, the LiDAR hardware is  arguably the most expensive additional component required for robust
autonomous driving. Without it, the additional hardware cost for autonomous driving becomes relatively minor.  Further,  image-based object detection would  also be  beneficial even in the presence of LiDAR equipment.  One could imagine a scenario where  the LiDAR data is used to continuously train and fine-tune an image-based classifier. In case of our sensor outage, the image-based classifier could likely function as a very reliable backup. Similarly, one could imagine a setting where high-end cars are shipped with LiDAR hardware and continuously train the image-based classifiers that are used in cheaper models.  

\paragraph{Future work.} There are multiple immediate directions along which our results could be improved in future work: First, higher resolution stereo images would likely  significantly improve the accuracy for faraway objects.  Our results were obtained with 0.4 megapixels ---  a far cry from the state-of-the-art camera technology. Second, in this paper we did not focus on real-time image processing  and the classification of all objects in one image takes on the order of 1s. However, it is likely possible to improve these speeds by several orders of magnitude. Recent improvements on real-time multi-resolution depth estimation~\cite{wang2018anytime} show that an effective way to speed up depth estimation is to first compute a depth map at low resolution and then incorporate high-resolution to refine the previous result. The conversion from a depth map to pseudo-LiDAR is very fast and it should be possible to  drastically speed up the detection pipeline through e.g. model distillation~\cite{bucilua2006model} or anytime prediction~\cite{huang2017multi}. Finally, it is likely that future work could improve the state-of-the-art in 3D object detection through sensor fusion of LiDAR and pseudo-LiDAR.   Pseudo-LiDAR  has the advantage that its signal is much denser than LiDAR and the two data modalities could have complementary strengths. 
We hope that our findings will cause a revival of image-based 3D object recognition and  our progress will motivate the computer vision community to fully close the image/LiDAR gap in the near future.

\section*{Acknowledgments}
This research is supported in part by grants from the National Science Foundation (III-1618134, III-1526012, IIS-1149882, IIS-1724282, and TRIPODS-1740822), the Office of Naval Research DOD (N00014-17-1-2175), and the Bill and Melinda Gates Foundation. We are thankful for generous support by Zillow and SAP America Inc. We thank Gao Huang for helpful discussion.

{\small
\bibliographystyle{ieee}
\bibliography{main}
}

\clearpage
\appendix
\begin{center}
	\textbf{\Large Supplementary Material}
\end{center}

In this Supplementary Material, we provide details omitted in the main text.
\begin{itemize}
	\item Section~\ref{sDetail}: additional details on our approach (Section~4.2 of the main paper).
	\item Section~\ref{sSPS}: results using \SPSstereo~\cite{yamaguchi2014efficient} (Section~4.3 of the main paper).
	\item Section~\ref{sDepth}: further analysis on depth estimation (Section~4.3 of the main paper).
	\item Section~\ref{sTest}: additional results on the test set (Section~4.4 of the main paper).
	\item Section~\ref{sQuali}: additional qualitative results (Section~4.5 of the main paper).
\end{itemize}
\section{Additional Details of Our Approach}
\label{sDetail}

\subsection{Ground plane estimation}
As mentioned in the main paper, \AVOD~\cite{ku2018joint} takes image-specific ground planes as inputs. A ground plane is parameterized by a normal vector $\vw=[w_x,w_y,w_z]^\top\in\mathbb{R}^3$ and a ground height $h\in\mathbb{R}$. We estimate the parameters according to the pseudo-LiDAR points $\{\vp^{(n)} = [x^{(n)}, y^{(n)}, z^{(n)}]^\top\}_{n=1}^N$ (see Section 3 of the main paper). Specifically, we consider points that  are close to the camera and fall into a certain range of possible ground heights:
\begin{align}
& \text{(width)} \hspace{20pt} 15.0 \geq x \geq -15.0,\\
& \text{(height)} \hspace{18pt} 1.86 \geq y \geq 1.5,\\
& \text{(depth)} \hspace{20pt} 40.0 \geq z \geq 0.0.
\end{align}
Ideally, all these points will be on the plane: $\vw^\top\vp+h = 0$. We fit the parameters with a straight-forward application of RANSAC~\cite{fischler1981random}, in which we constraint $w_y=-1$. We then normalize the resulting $\vw$ to have a unit $\ell_2$ norm.

\subsection{Pseudo disparity ground truth}
We train a version of \PSMNet~\cite{chang2018pyramid} (named \PSMNetpd) using the 3,712 training images of detection, instead of the 200 KITTI stereo images~\cite{geiger2012we,menze2015object}.
We obtain pseudo disparity ground truth as follows: We project the corresponding LiDAR points into the 2D image space, followed by applying Eq. (1) of the main paper to derive disparity from pixel depth. If multiple LiDAR points are projected to a single pixel location, we randomly keep one of them. We ignore those pixels with no depth (disparity) in training \PSMNet.

\section{Results Using \SPSstereo~\cite{yamaguchi2014efficient}}
\label{sSPS}

\begin{table}
	\centering
	\caption{\small Comparison of different stereo disparity methods on pseudo-LiDAR-based detection accuracy with \AVOD. We report \APBEV ~/ \AP (in \%) of the \textbf{moderate car} category at IoU = 0.7.} \label{tbSPS}

	\begin{tabular}{c|l|c}
		Method & \multicolumn{1}{c|}{Disparity} & \APBEV ~/ \AP \\ \hline
& \SPSstereo & 39.1 / 28.3 \\
& \DispNetS & 36.3 / 27.0  \\
\AVOD & \DispNetC  & 36.5 / 26.2 \\
& \PSMNet  & 39.2 / 27.4 \\
& \PSMNetpd  & 56.8 / 45.3  \\
		\hline
	\end{tabular}
\end{table}

In Table~\ref{tbSPS}, we report the 3D object detection accuracy of pseudo-LiDAR with \SPSstereo~\cite{yamaguchi2014efficient}, a non-learning-based stereo disparity approach. On the leaderboard of KITTI stereo 2015, \SPSstereo achieves 3.84\% disparity error, which is worse than the error of 1.86\% by \PSMNet but better than 4.32\% by \DispNetC. The object detection results with \SPSstereo are on par with those with \PSMNet and \DispNet, even if it is not learning-based.

\section{Further Analysis on Depth Estimation}
\label{sDepth}

We study how over-smoothing the depth estimates would impact the 3D object detection accuracy. We train \AVOD~\cite{ku2018joint} and \Frustum~\cite{qi2018frustum} using pseudo-LiDAR with \PSMNetpd. During evaluation, we obtain over-smoothed depth estimates using an average kernel of size $11\times 11$ on the depth map. Table~\ref{t_smooth} shows the results: over-smoothing leads to degraded performance, suggesting the importance of high quality depth estimation for accurate 3D object detection. 

\begin{table}[t]
	\centering
	\small
	\caption{\small The impact of over-smoothing the depth estimates on the 3D detection results. We evaluate pseudo-LiDAR with \PSMNetpd. We report \APBEV ~/ \AP (in \%) of the \textbf{moderate car} category at IoU = 0.7 on the validation set.}
	\begin{tabular}{l|c|c}
	\multicolumn{1}{c|}{}  & \multicolumn{2}{c}{Detection algorithm} \\ \cline{2-3}
	\multicolumn{1}{c|}{Depth estimates} & \AVODGT & \Frustum \\ \hline
	Non-smoothed & 56.8 / 45.3 & 51.8 / 39.8 \\ \hline
	Over-smoothed & 53.7 / 37.8 & 48.3 / 31.6 \\ \hline
	\end{tabular}
	\label{t_smooth}
\end{table} 

\section{Additional Results on the Test Set}
\label{sTest}
We report the results on the pedestrian and cyclist categories on the KITTI test set in Table~\ref{t_test_pedestrian}. For \Frustum which takes 2D bounding boxes as inputs, \cite{qi2018frustum} does not provide the 2D object detector trained on KITTI or the detected 2D boxes on the test images. Therefore, for the car category we apply the released RRC detector~\cite{ren2017accurate} trained on KITTI (see Table~\ref{tbTest} in the main paper). For the pedestrian and cyclist categories, we apply Mask R-CNN~\cite{he2017mask} trained on MS COCO~\cite{lin2014microsoft}. The detected 2D boxes are then inputted into \Frustum~\cite{qi2018frustum}.
We note that, MS COCO has no cyclist category. We thus use the detection results of bicycles as the substitute.

On the pedestrian category, we see a similar gap between pseudo-LiDAR and LiDAR as the validation set (cf. Table~\ref{tbPedestrian} in the main paper). However, on the cyclist category we see a drastic performance drop by pseudo-LiDAR. This is likely due to the fact that cyclists are relatively uncommon in the KITTI dataset and the algorithms have over-fitted. For \Frustum, the detected bicycles may not provide accurate heights for cyclists, which essentially include riders and bicycles. Besides, the detected bicycles without riders are false positives to cyclists, hence leading to a much worse accuracy. 

\emph{We note that, so far no image-based algorithms report 3D results on these two categories on the test set.}

\begin{table}
	\centering
	\small
	\tabcolsep 2pt
	\caption{\small 3D object detection results on the \textbf{pedestrian} and \textbf{cyclist} categories on the \emph{test} set. We compare pseudo-LiDAR with \PSMNetpd (in {\color{blue}blue}) and LiDAR (in {\color{gray} gray}). We report \APBEV~/ \AP at IoU = 0.5 (the standard metric). $\dagger$: Results on the KITTI leaderboard.} \label{t_test_pedestrian}
	\resizebox{\linewidth}{!}{  
	\begin{tabular}{=c|+c|+c|+c|+c}
		Method & Input signal  & Easy & Moderate & Hard \\ \hline
		\multicolumn{5}{c}{Pedestrian} \\ \hline
		\rowstyle{\color{blue}}
		\AVOD & Stereo & 27.5 / 25.2 & 20.6 / 19.0 & 19.4 / 15.3\\
		\rowstyle{\color{blue}}
		\Frustum & Stereo &  31.3 / 29.8 &  24.0 / 22.1  &  21.9 / 18.8 \\
		\hline
		\rowstyle{\color{gray}}
		\AVOD & $\dagger$LiDAR + Mono & 58.8 / 50.8 & 51.1 / 42.8 & 47.5 / 40.9 \\
		\rowstyle{\color{gray}}
		\Frustum & $\dagger$LiDAR + Mono & 58.1 / 51.2 & 50.2 / 44.9 & 47.2 / 40.2 \\ \hline
		\multicolumn{5}{c}{Cyclist} \\ \hline
		\rowstyle{\color{blue}}
		\AVOD & Stereo & 13.5 / 13.3 & 9.1 / 9.1 & 9.1 / 9.1\\
		\rowstyle{\color{blue}}
		\Frustum & Stereo &  4.1 / 3.7 &  3.1 / 2.8 & 2.8 / 2.1\\
		 \hline
		\rowstyle{\color{gray}}
		\AVOD & $\dagger$LiDAR + Mono & 68.1 / 64.0 & 57.5 / 52.2 & 50.8 / 46.6 \\
		\rowstyle{\color{gray}}
		\Frustum & $\dagger$LiDAR + Mono & 75.4 / 72.0 & 62.0 / 56.8 & 54.7 / 50.4 \\
	\end{tabular}
	}
\end{table}

\section{Additional Qualitative Results}
\label{sQuali}

\begin{figure}[t]
	\centerline{\includegraphics[width=\linewidth]{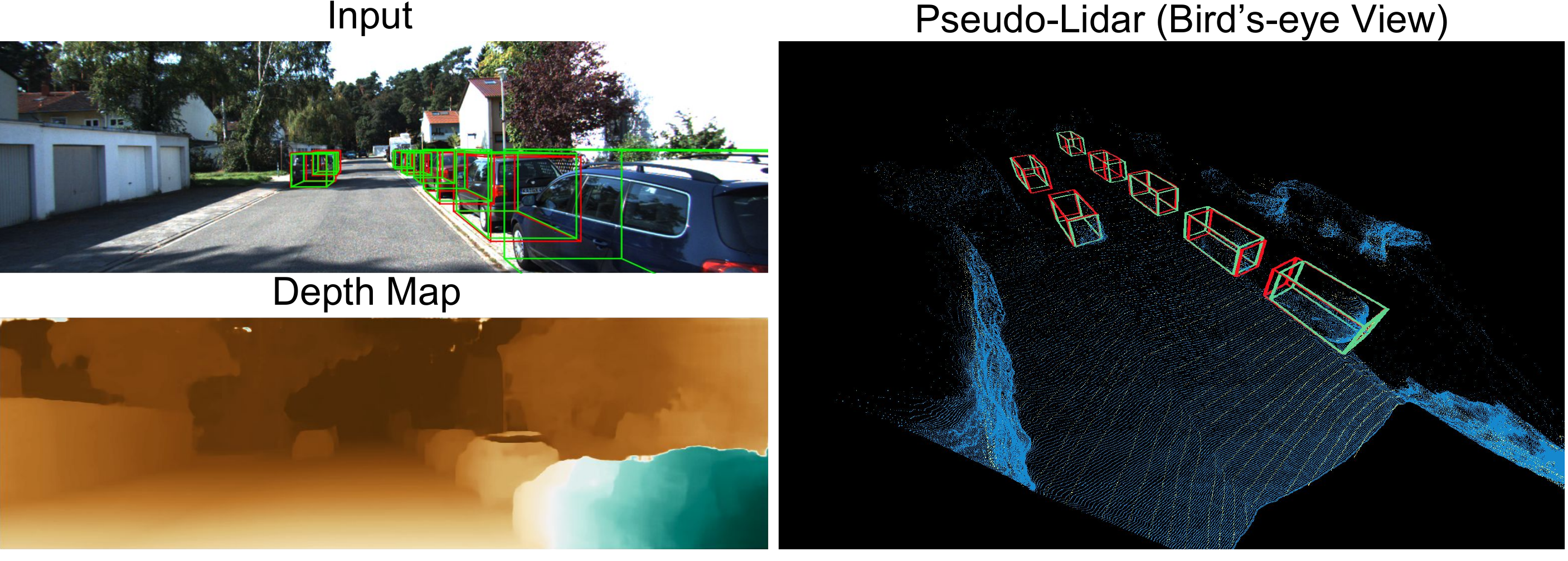}}
	\caption{\textbf{Pseudo-LiDAR signal from visual depth estimation.} Top-left: a KITTI street scene with super-imposed bounding boxes around cars obtained with LiDAR ({\color{red}red}) and pseudo-LiDAR ({\color{green}green}). Bottom-left: estimated disparity map. Right: pseudo-LiDAR ({\color{blue}blue}) vs. LiDAR ({\color{yellow}yellow}) --- the pseudo-LiDAR points align remarkably well with the LiDAR ones. Best viewed in color (zoom in for details).}
	\label{fig:stere_lidar_suppl}
\end{figure}

\subsection{LiDAR vs. pseudo-LiDAR}
We include in Fig.~\ref{fig:stere_lidar_suppl} more qualitative results comparing the LiDAR and pseudo-LiDAR signals. The pseudo-LiDAR points are generated by \PSMNetpd. Similar to Fig.~\ref{fig:stere_lidar} in the main paper, the two modalities align very well. 

\subsection{\textbf{\PSMNet} vs. \textbf{\PSMNetpd}}
\begin{figure*}[t]
	\centerline{\includegraphics[width=0.8\linewidth]{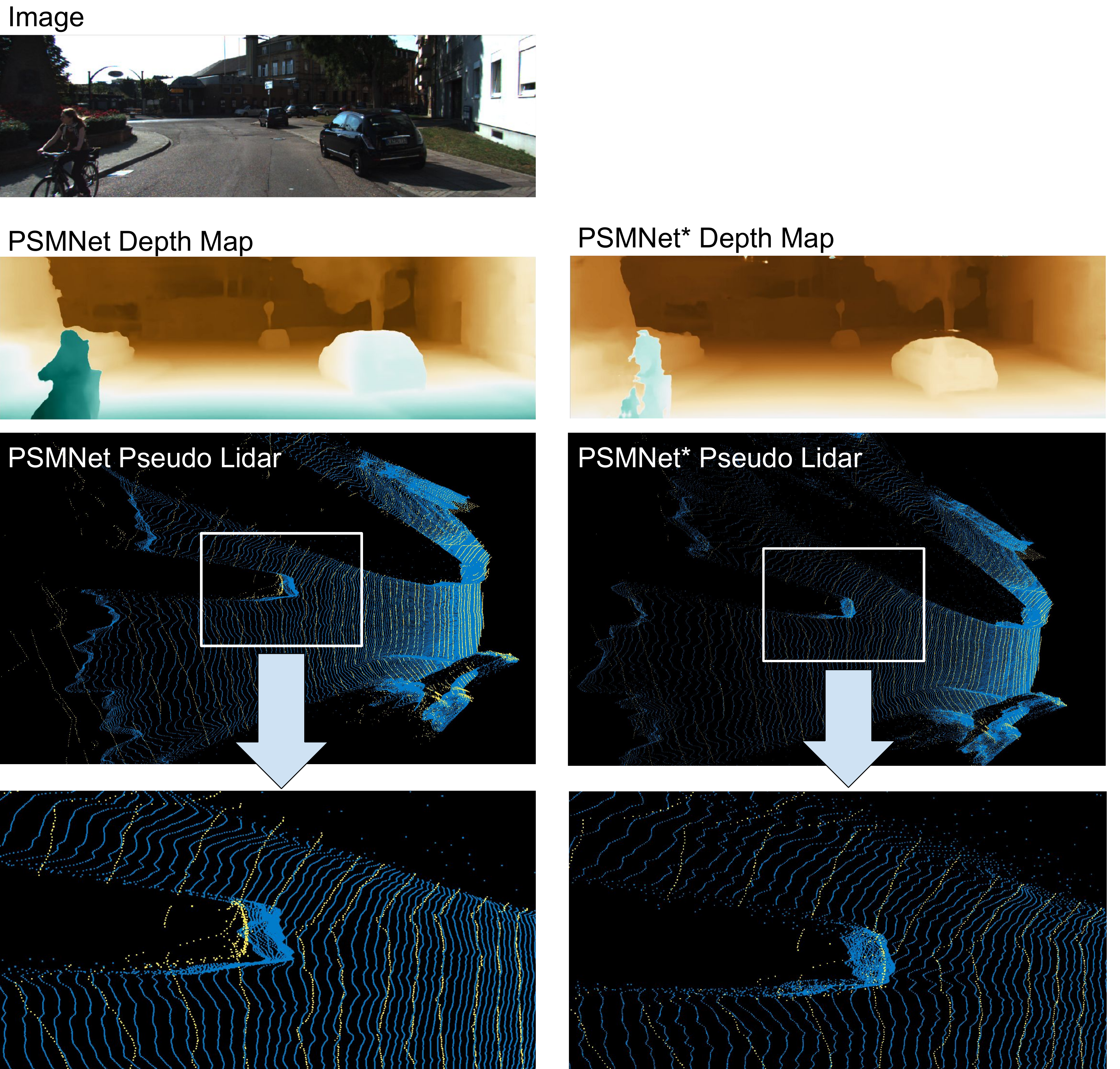}}
	\caption{\textbf{\PSMNet vs. \PSMNetpd.} Top: a KITTI street scene. Left column: the depth map and pseudo-LiDAR points (from the bird's-eye view) by \PSMNet, together with a zoomed-in region.  Right column: the corresponding results by \PSMNetpd. The observer is on the very right side looking to the left. The pseudo-LiDAR points are in {\color{blue}blue}; LiDAR points are in {\color{yellow}yellow}. The pseudo-LiDAR points by \PSMNet have larger deviation at far-away distances. Best viewed in color (zoom in for details).}
	\label{fig:psmnet_suppl}
\end{figure*}

We further compare the pseudo-LiDAR points generated by \PSMNetpd and \PSMNet. The later is trained on the 200 KITTI stereo images with provided denser ground truths. As shown in Fig.~\ref{fig:psmnet_suppl}, the two models perform fairly similarly for nearby distances. For far-away distances, however, the pseudo-LiDAR points by \PSMNet start to show notable deviation from LiDAR signal. This result suggest that significant further
improvements could be possible through learning disparity on a large training set or even end-to-end training of the whole pipeline.

\subsection{Visualization and failure cases}
We provide additional visualization of the prediction results (cf. Section 4.5 of the main paper). We consider \AVOD with the following point clouds and representations.
\begin{itemize}
	\item LiDAR
	\item pseudo-LiDAR (stereo): with \PSMNetpd~\cite{chang2018pyramid}
	\item pseudo-LiDAR (mono): with \DORN~\cite{fu2018deep}
	\item frontal-view (stereo): with \PSMNetpd~\cite{chang2018pyramid}
\end{itemize}
We note that, as \DORN~\cite{fu2018deep} applies ordinal regression, the predicted monocular depth are discretized.
 
\begin{figure*}[t]
	\centerline{\includegraphics[width=0.8\linewidth]{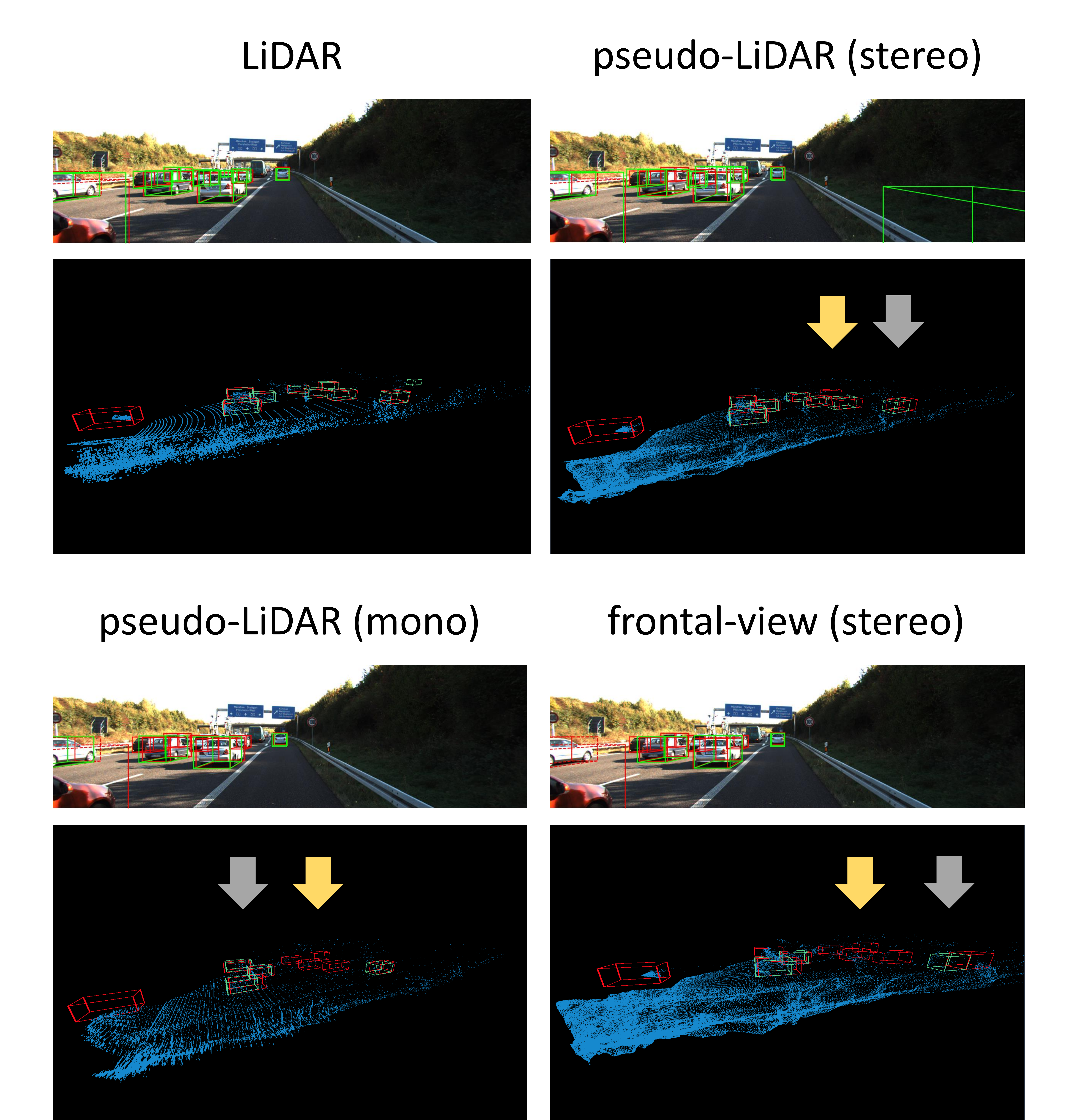}}
	\caption{\textbf{Qualitative comparison and failure cases.} We compare \AVOD with LiDAR, pseudo-LiDAR (stereo), pseudo-LiDAR (monocular), and frontal-view (stereo). Ground-truth boxes are in {\color{red}{red}}; predicted boxes in {\color{green}{green}}. The observer in the pseudo-LiDAR plots (bottom row) is on the very left side looking to the right. The \textbf{mislocalization} cases are indicated by {\color{gray}gray} arrows; the \textbf{missed detection} cases are indicated by {\color{yellow}yellow} arrows. The frontal-view approach (\emph{bottom-right}) makes extremely inaccurate predictions, even for nearby objects. Best viewed in color.}
	\label{fig:qualitative_suppl}
\end{figure*}

As shown in Fig.~\ref{fig:qualitative_suppl}, both LiDAR and pseudo-LiDAR (stereo or mono) lead to accurate predictions for the nearby objects. However, pseudo-LiDAR 
detects far-away objects less precisely (\textbf{mislocalization}: {\color{gray}gray} arrows) or even fails to detect them (\textbf{missed detection}: {\color{yellow}yellow} arrows) due to in-accurate depth estimates, especially for the monocular depth. For example, pseudo-LiDAR (mono) completely misses the four cars in the middle. On the other hand, the frontal-view (stereo) based approach makes extremely inaccurate predictions, even for nearby objects.

\begin{figure*}
	\centering
	\begin{subfigure}[b]{0.45\textwidth}
		\includegraphics[width=\textwidth]{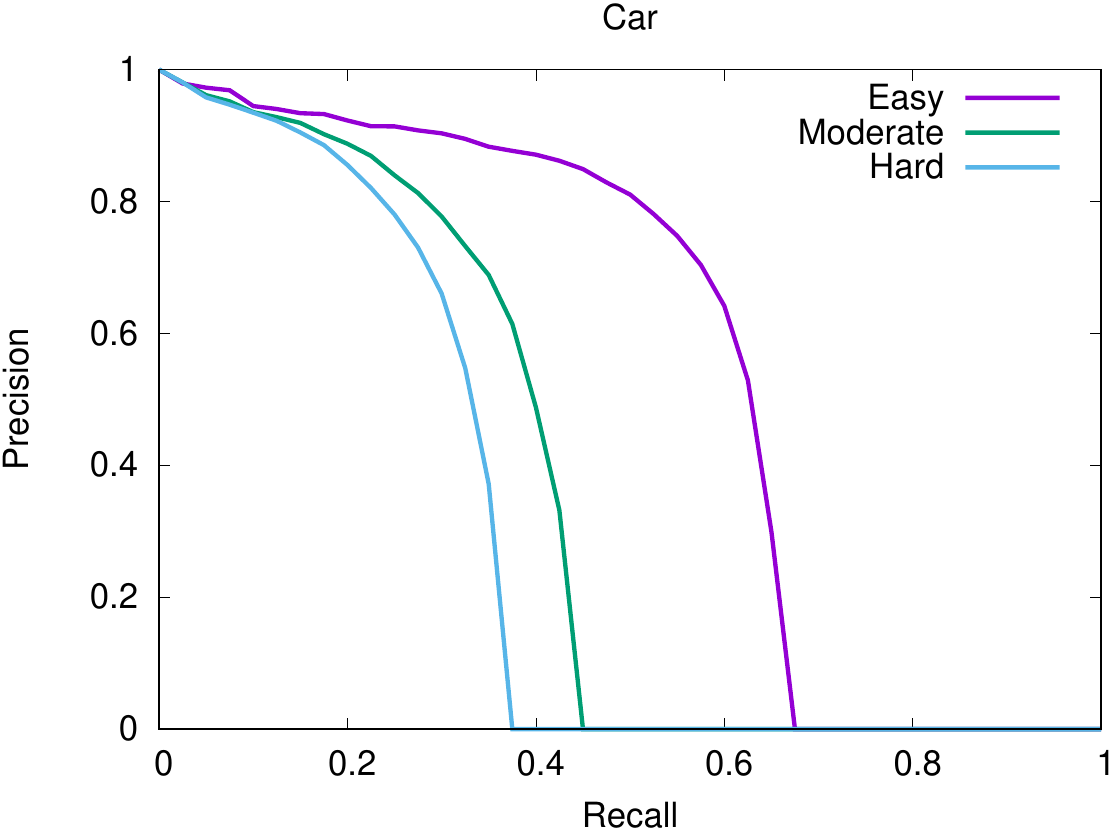}
		\subcaption{3D detection: \AVOD + pseudo-LiDAR (stereo)}
	\end{subfigure}
	\begin{subfigure}[b]{0.45\textwidth}
		\includegraphics[width=\textwidth]{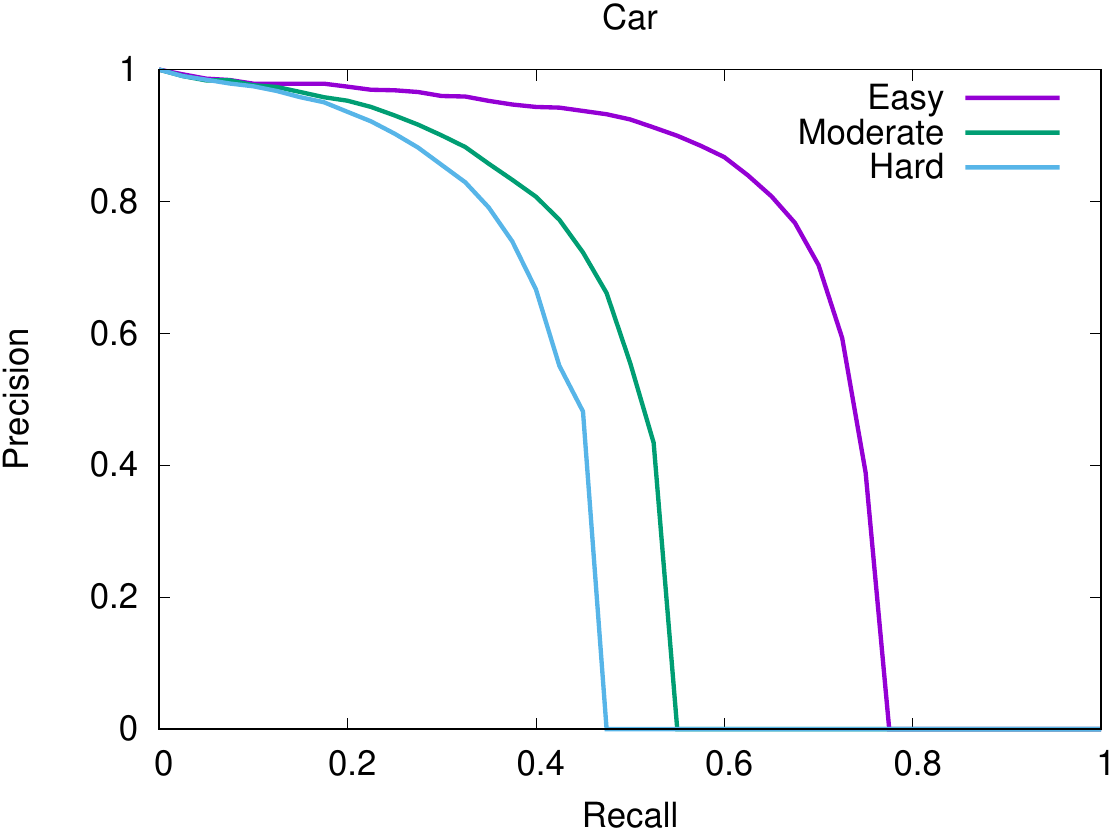}
		\subcaption{BEV detection: \AVOD + pseudo-LiDAR (stereo)}
	\end{subfigure}
	\vskip 20pt
	\begin{subfigure}[b]{0.45\textwidth}
		\includegraphics[width=\textwidth]{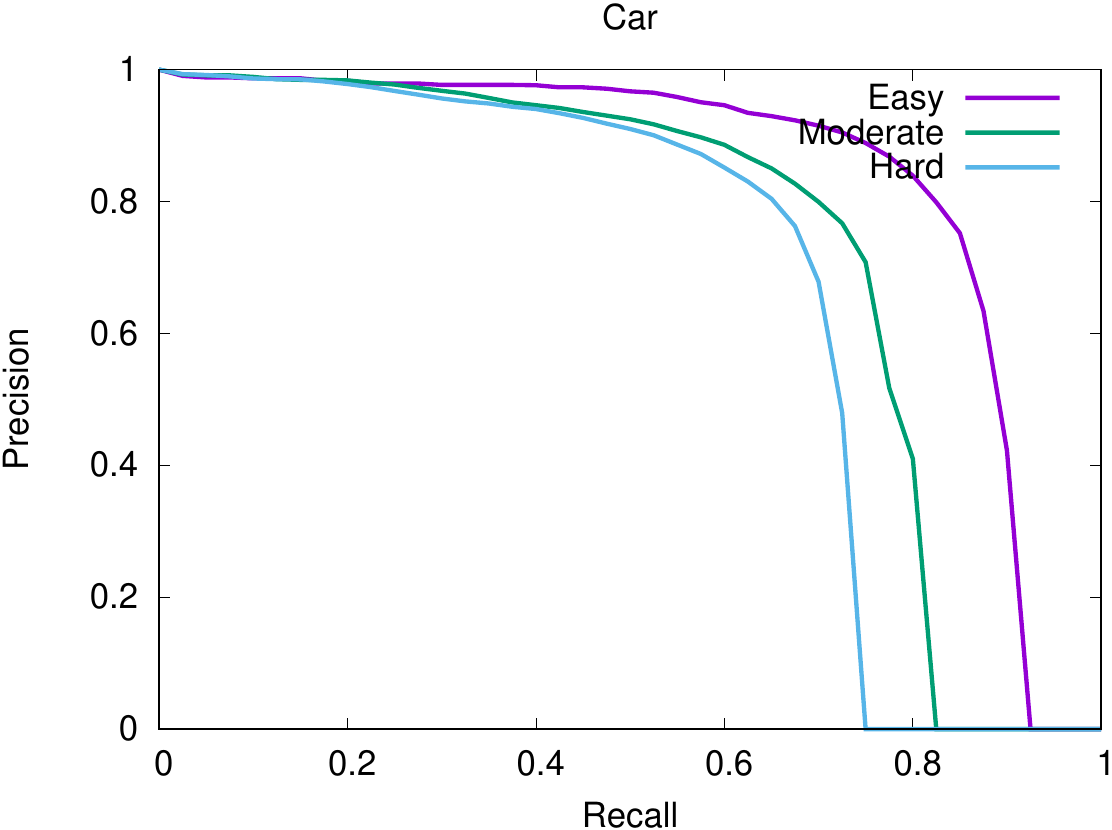}
		\subcaption{3D detection: \AVOD + LiDAR}
	\end{subfigure}
	\begin{subfigure}[b]{0.45\textwidth}
		\includegraphics[width=\textwidth]{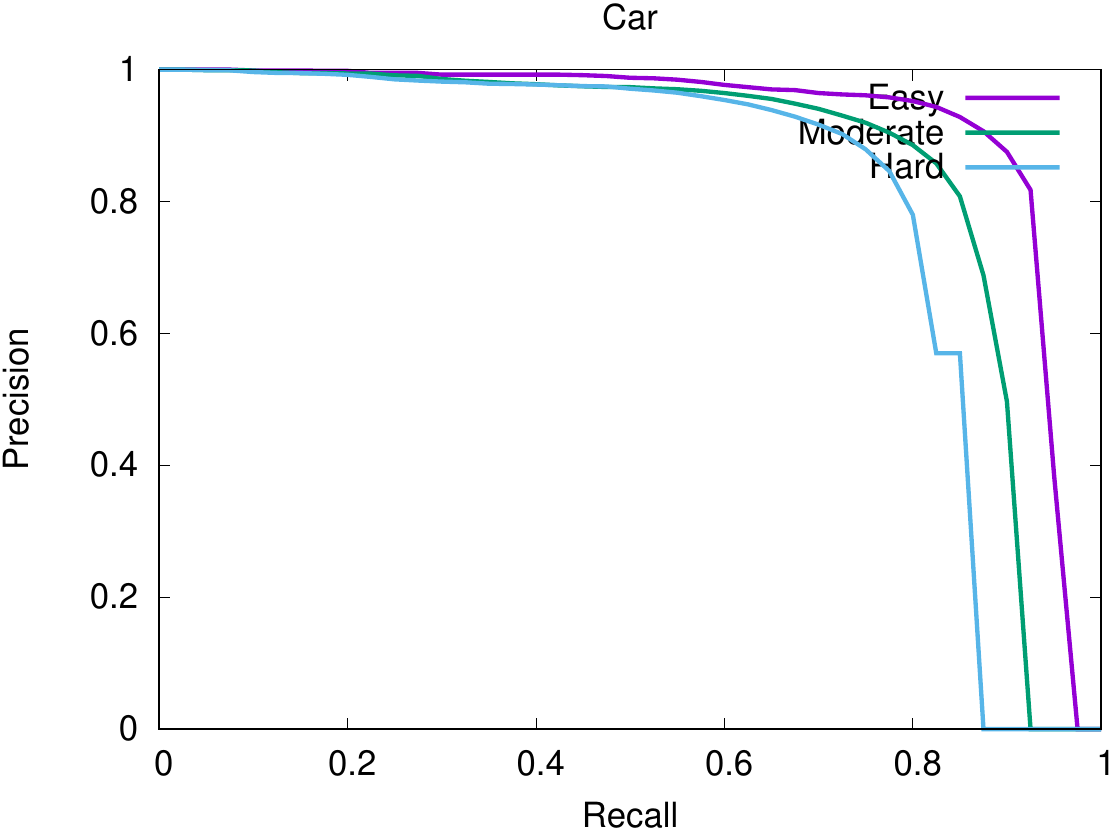}
		\subcaption{BEV detection: \AVOD + LiDAR}
	\end{subfigure}		
	\caption{\textbf{Precision-recall curves.} We compare the precision and recall on \AVOD using pseudo-LiDAR with \PSMNetpd (top) and using LiDAR (bottom) on the test set. We obtain the curves from the KITTI website. We show both the 3D detection results (left) and the BEV detection results (right). \AVOD using pseudo-LiDAR has a much lower recall, suggesting that missed detections are one of the major issues of pseudo-LiDAR-based detection.}
	\label{fig:PR_curve}
\end{figure*}

To analyze the failure cases, we show the precision-recall (PR) curves on both 3D object and BEV detection in Fig.~\ref{fig:PR_curve}. The pseudo-LiDAR-based detection has a much lower recall compared to the LiDAR-based one, especially for the moderate and hard cases (i.e., far-away or occluded objects). That is, missed detections are one major issue that pseudo-LiDAR-based detection needs to resolve.

\begin{figure*}[t]
	\centerline{\includegraphics[width=0.8\linewidth]{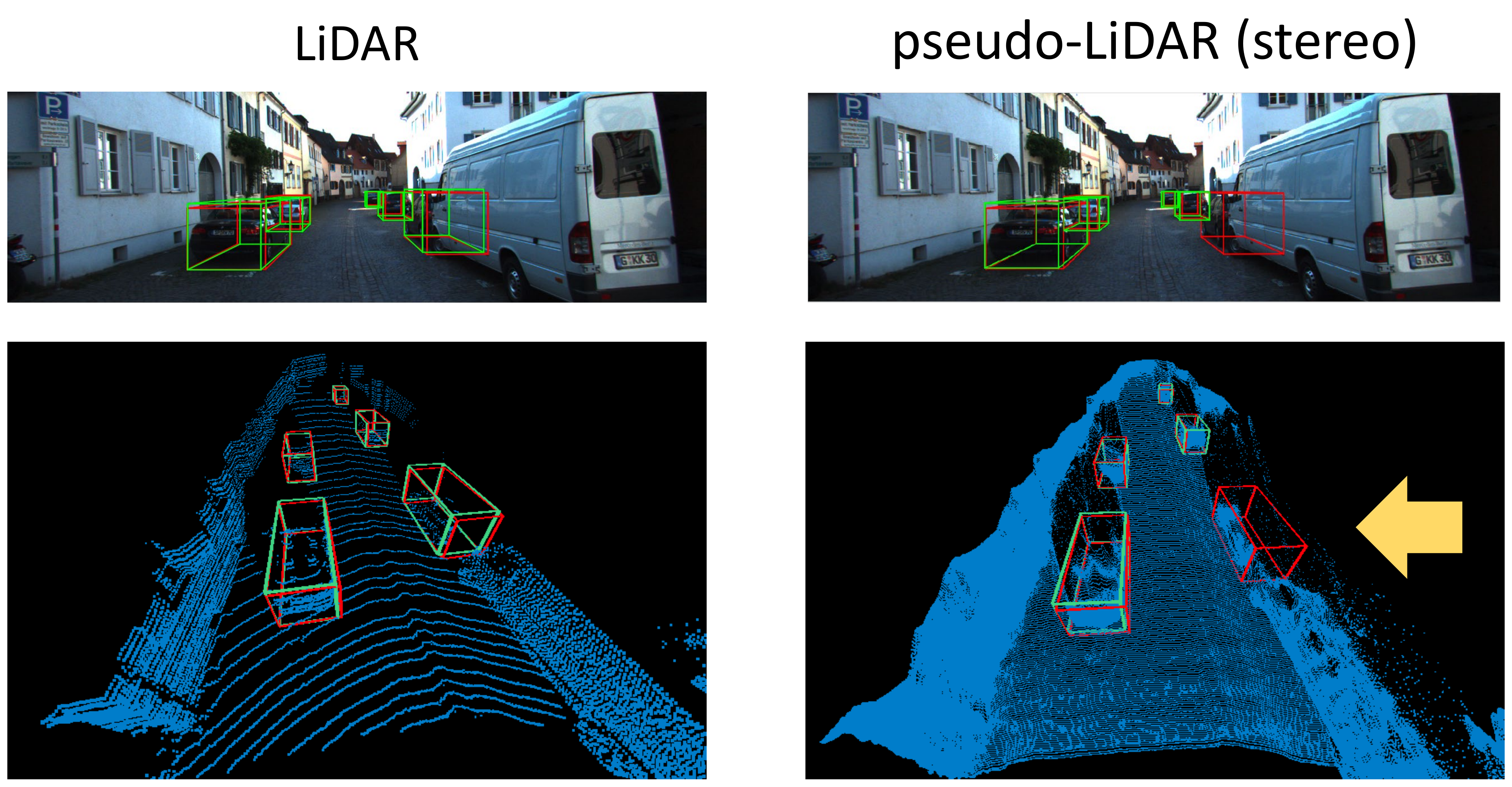}}
	\caption{\textbf{Qualitative comparison and failure cases.} We compare \AVOD with LiDAR and pseudo-LiDAR (stereo). Ground-truth boxes are in {\color{red}{red}}; predicted boxes in {\color{green}{green}}. The observer in the pseudo-LiDAR plots (bottom row) is on the bottom side looking to the top. The pseudo-LiDAR-based detection misses the partially occluded car (the {\color{yellow} yellow} arrow), which is a hard case. Best viewed in color.}
	\label{fig:faliure_suppl}
\end{figure*}

We provide another qualitative result for failure cases in Fig.~\ref{fig:faliure_suppl}. The partially occluded car is missed detected by \AVOD with pseudo-LiDAR (the {\color{yellow} yellow} arrow) even if it is close to the observer, which likely indicates that stereo disparity approaches suffer from noisy estimation around occlusion boundaries.

\end{document}